\def\year{2022}\relax
\documentclass[letterpaper]{article} % DO NOT CHANGE THIS
\usepackage{aaai22}  % DO NOT CHANGE THIS
\usepackage{times}  % DO NOT CHANGE THIS
\usepackage{helvet}  % DO NOT CHANGE THIS
\usepackage{courier}  % DO NOT CHANGE THIS
\usepackage[hyphens]{url}  % DO NOT CHANGE THIS
\usepackage{graphicx} % DO NOT CHANGE THIS
\usepackage{natbib}  % DO NOT CHANGE THIS AND DO NOT ADD ANY OPTIONS TO IT
\usepackage{caption} % DO NOT CHANGE THIS AND DO NOT ADD ANY OPTIONS TO IT
\DeclareCaptionStyle{ruled}{labelfont=normalfont,labelsep=colon,strut=off} % DO NOT CHANGE THIS
\usepackage{algorithm}
\usepackage{algorithmic}

\usepackage{newfloat}
\usepackage{listings}
\floatstyle{ruled}
\newfloat{listing}{tb}{lst}{}
\floatname{listing}{Listing}
\usepackage{booktabs}
\usepackage{multirow}
\usepackage{makecell}
\usepackage{amssymb}
\usepackage{bbding}

\title{Local Similarity Pattern and Cost Self-Reassembling for Deep Stereo Matching Networks}
\author{
        Biyang Liu \textsuperscript{\rm 1,2},
        Huimin Yu \textsuperscript{\rm 1,2,3,4},
        Yangqi Long \textsuperscript{\rm 1, 2} \\
}
\affiliations{
    \textsuperscript{\rm 1} College of Information Science and Electronic Engineering, Zhejiang University \\ 
    \textsuperscript{\rm 2} ZJU-League Research \& Development Center,   
    \textsuperscript{\rm 3} State Key Lab of CAD\&CG, Zhejiang University \\
    \textsuperscript{\rm 4} Zhejiang Provincial Key Laboratory of Information Processing, Communication and Networking \\
    \{biyangliu, yhm2005, longyangqi\}@zju.edu.cn
}

\begin{document}

\maketitle

\begin{abstract}
Although convolution neural network based stereo matching architectures have made impressive achievements, there are still some limitations: 1) Convolutional Feature (CF) tends to capture appearance information, which is inadequate for accurate matching. 2) Due to the static filters, current convolution based disparity refinement modules often produce over-smooth results. In this paper, we present two schemes to address these issues, where some traditional wisdoms are integrated. Firstly, we introduce a pairwise feature for deep stereo matching networks, named LSP (Local Similarity Pattern). Through explicitly revealing the neighbor relationships, LSP contains rich structural information, which can be leveraged to aid CF for more discriminative feature description. Secondly, we design a dynamic self-reassembling refinement strategy and apply it to the cost distribution and the disparity map respectively. The former could be equipped with the unimodal distribution constraint to alleviate the over-smoothing problem, and the latter is more practical. The effectiveness of the proposed methods is demonstrated via incorporating them into two well-known basic architectures, GwcNet and GANet-deep. Experimental results on the SceneFlow and KITTI benchmarks show that our modules significantly improve the performance of the model.
\end{abstract}

\section{Introduction}
Depth information is essential for plenty of applications such as autonomous driving and augmented reality. As a convenient and low-cost means to obtain depth, stereo matching has been studied for a long time. The objective of this task is to estimate the disparity from a rectified image pair, which can be converted to depth with the focal length and baseline of the camera system. Typically, stereo matching algorithms consist of four steps \cite{Scharstein2002A}: matching cost computation, cost aggregation, disparity optimization, and disparity refinement.

Although convolutional neural network (CNN) based stereo matching methods have been dominant on several benchmarks, the vanilla convolution is not perfect, there are still some flaws in CNN based architectures. Previous works have proved that integrating traditional knowledge will benefit the designing of the model \cite{zhang2019ga, xu2020aanet}. However, they concentrate more on the cost aggregation stage. In this paper, we mainly investigate some limitations of the used features and the refinement modules. Aiming at breaking the limitations, we present two schemes, introduced as follows.

\begin{figure}[t]
\centering
\includegraphics[width=0.9\columnwidth]{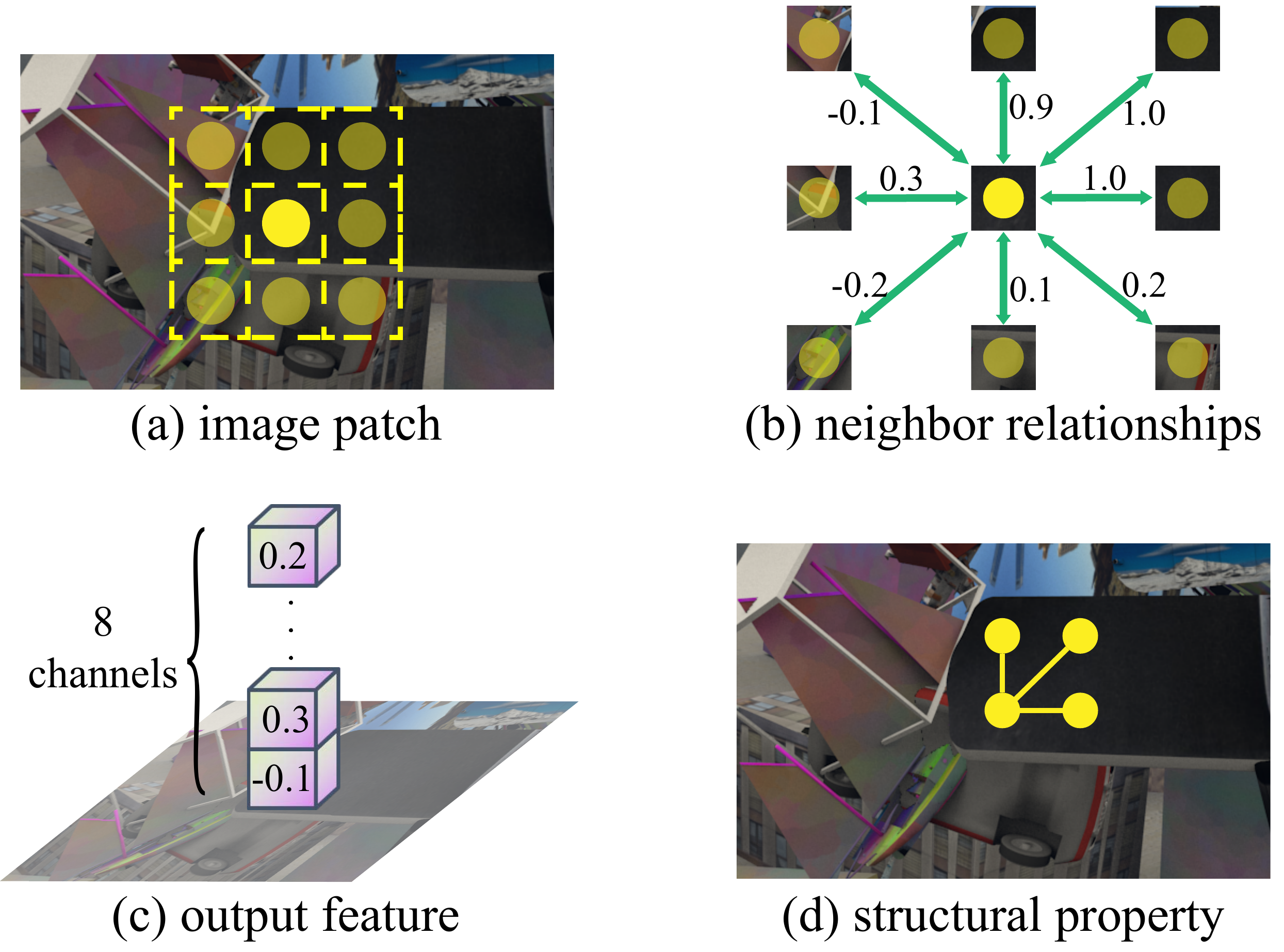} 
\caption{The operation paradigm of LSP. Each yellow circle represents a point in the image or the feature map. Local structural property is reflected in the close relationships.}
\label{Fig1}
\end{figure}

1) Convolutional Feature (CF) is utilized to describe the points to construct the matching cost in current stereo matching networks. Despite the rich appearance information, CF contains less structural information since the learned convolutional kernels can not explicitly reveal the pixel relationships. Due to the view transformation, the appearance of an object might get changed while the structural characteristic is more stable \cite{zabih1994non}. Therefore, it is essential for accurate matching to combine the appearance information and the structural information, as demonstrated in AD-Census \cite{mei2011building}.

To this end, we design a pairwise feature for deep stereo matching networks, named LSP (Local Similarity Pattern). Based on the unary features extracted from the convolutional layers, the relationships (e.g. cosine similarities) of a point and its neighbors are calculated as LSP, as illustrated in Figure \ref{Fig1}. With such an operating mechanism, LSP clearly discloses some structural clues (e.g. trees are vertical), making it a beneficial complement for CF. At the same time, LSP is calculated in a bottom-up fashion \cite{hu2019local}, thus it is unnecessary to iterate multiple times to obtain a representative description. Further to make LSP more complete, we extend it with a multi-scale and multi-level scheme. The former is aimed at handling objects of various sizes, and the latter is designed considering the unary features from different convolutional layers contain different information.

2) At present, most refinement methods are based on a CNN architecture that takes multi-modal inputs to predict a disparity residual map \cite{khamis2018stereonet, liang2018learning}. We observe the results from these methods tend to be over-smooth \cite{chen2019over}. The reason might be that the learned static and shared convolutional filters are difficult to restore various error patterns. Under the guidance of the $L1$ loss, convolutions try to reduce the average error. It is important to address the over-smoothing problem, especially for some downstream tasks, such as 3D object detection \cite{garg2020wasserstein}.

Inspired by the elegant disparity filling methods \cite{huq2013occlusion}, we propose a dynamic refinement strategy CSR (Cost Self-Reassembling). The basic hypothesis is that the true cost distribution of an outlier should be close to those of some neighbors. Rather than manually searching for these close and reliable neighbors, we build a network and force it to learn to predict the locations of the neighbors by itself. In this way, the disparity propagation process will adapt to the image content and is not limited to the regular grid, making CSR more powerful than other CNN based refinement modules. By further imposing the unimodal distribution constraint on the refined cost, the over-smoothing problem can be effectively mitigated \cite{chen2019over}. In addition to the cost distribution, the dynamic reassembling strategy could be applied to the disparity map and DSR (Disparity Self-Reassembling) is more suitable for practical algorithms.

Our main contributions are summarized as:
\begin{itemize}
\item We introduce a pairwise feature LSP for deep stereo matching networks. The structural information contained in LSP makes it a beneficial supplement for CF.
\item We propose two dynamic disparity refinement strategies CSR and DSR. In particular, CSR can be equipped with the unimodal distribution constraint to alleviate the over-smoothing problem.
\item Through inserting the designed modules, we surpass the baseline GwcNet by a substantial margin on the SceneFlow and KITTI benchmarks. We also incorporate them into a state-of-the-art (SOTA) model GANet-deep and achieve a significant performance improvement.
\end{itemize}

\section{Related Works}

\subsection{Deep Stereo Matching Networks}
In early works, CNNs were used to compute the matching costs \cite{zbontar2015computing, chen2015deep, luo2016efficient}, followed by traditional cost aggregation and disparity refinement methods. Nowadays, end-to-end frameworks are more popular. According to the dimension of the convolutions used in cost aggregation, they can be divided into two categories: 2D-CNN based models \cite{mayer2016large, song2020edgestereo, xu2020aanet} and 3D-CNN based models \cite{kendall2017end, chang2018pyramid, Guo_2019_CVPR, zhang2019ga}. Generally speaking, stereo matching networks belonging to the former category are more efficient, yet those belonging to the latter category perform better. 

\subsection{Structural Features}
CT (Census Transform) \cite{zabih1994non} and LBP (Local Binary Pattern) \cite{ojala2002multiresolution} are two well-known structural features in the field of computer vision. Despite the great achievements made in traditional algorithms, the non-differentiable Heaviside step function prevents them from being applied to the learned features of CNNs. In LSP, we alternatively utilize cosine similarity to measure the relationships. Apart from making LSP trainable, it solves the problem that the input of CT can only be the 1-channel grayscale image. We also note the affinity feature in \cite{gan2018monocular}. Except for LSP is more complete with a multi-scale and multi-level scheme, our motivation is different. Affinity feature is used to understand the depth relationships \cite{gan2018monocular}, while LSP is to describe a point with structural information.

\subsection{Disparity Refinement}
In current refinement modules, a CNN architecture is built to predict the disparity residual given the image features, projection error, etc \cite{khamis2018stereonet, liang2018learning}. Due to the stationarity of the convolutional filters, existing methods tend to produce over-smooth results \cite{chen2019over}. The problem is alleviated by our CSR, which is inspired by the classic disparity filling methods \cite{huq2013occlusion, tosi2019leveraging}. At the same time, it absorbs the advantage of deep learning. Instead of manually searching, we predict the locations of the neighbors with a network \cite{dai2017deformable} based on the image content. A similar strategy is adopted in the refinement of the depth maps \cite{ramamonjisoa2020predicting}. From the formula, it can be regarded as an instance of our DSR that assembles only one neighbor. Moreover, we predict the offsets without the initial disparity since our module can learn to find accurate neighbors by itself.

\section{Method}

\subsection{Overall Architecture}

\begin{figure*}[t]
\centering
\includegraphics[width=0.8\textwidth]{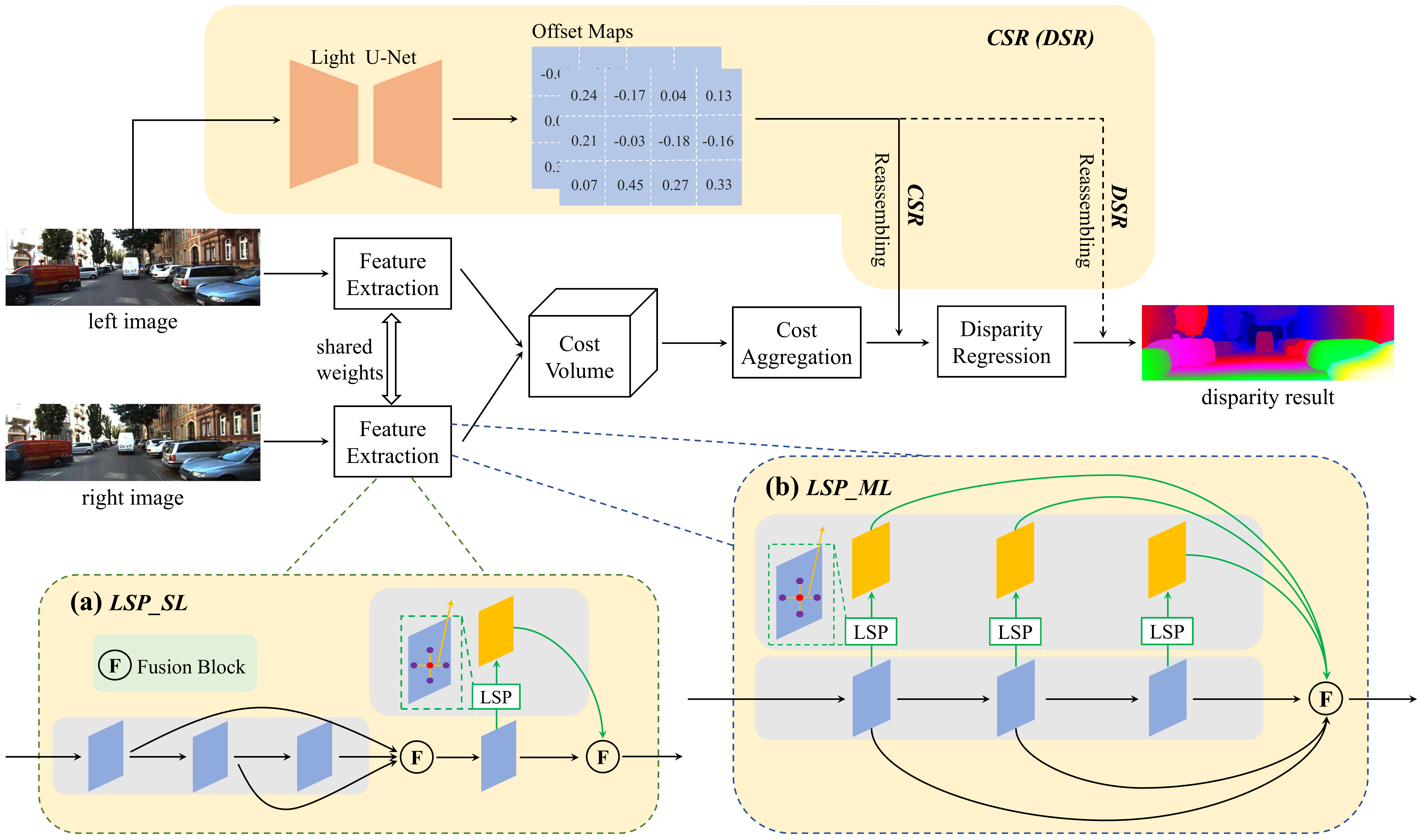} 
\caption{Overall architecture of our model based on GwcNet. Single-level or multi-level LSP is calculated with the convolutional features, then they are fused together for subsequent processings. The cost volume is formed via concatenating the left and the right features and is further aggregated by several 3D convolutions. Final disparity result is regressed by \emph{softmax} and weighted average. The self-reassembling refinement strategy can be applied to the cost distribution or the disparity map.}
\label{model}
\end{figure*}

In the following sections, we will describe the proposed modules and show how to insert them into a general deep stereo matching network. With GwcNet \cite{Guo_2019_CVPR} as the basic model, the overall architecture is shown in Figure \ref{model}, where the feature extraction module after embedding single-level or multi-level LSP and the refinement module are detailedly displayed. Note the cost volume in our model is constructed by feature concatenation rather than group-wise correlation \cite{Guo_2019_CVPR} since feature concatenation is a more popular manner.

\subsection{Local Similarity Pattern}

While CF contains rich appearance information, the structural information is often ignored. The reason lies in the fact that the learned weights in convolution can not explicitly reveal the neighbor relationships. To compensate for the lack, we introduce a pairwise feature, referred to as LSP. In LSP, each channel represents the relationship between the central point and one of its neighbors. It is formulated as: 
\begin{equation}
f_L^k(x, y) = \Phi(\mathbf{f}_C(x, y), \mathbf{f}_C(x+\delta x_k, y+\delta y_k))
\end{equation}
where $f_L^k(x, y)$ denotes the $k$-th channel of the LSP of point $(x, y)$, $\mathbf{f}_C$ is the convolutional feature with multiple channels, and $(\delta x_k, \delta y_k)$ represents the offset of the $k$-th neighbor point. $\Phi(\cdot, \cdot)$ is a function to measure the similarity of two vectors,  which will output a scalar. We select cosine similarity for its numerical stability.

Similar to convolution, we adopt a square window to define the neighborhood, as illustrated in Figure \ref{Fig1}. Note the output channels of $\mathbf{f}_L$ are only decided by the number of the neighbor points, i.e. the size of the window. Another important difference from CF is that LSP is dynamic and manually calculated \cite{hu2019local}. By this means, it is unnecessary to stack multiple layers to recognize various local patterns, making LSP convenient to embed.

Intuitively, smaller neighborhoods are beneficial for the matching at disparity discontinuities, while larger neighborhoods can help to understand the structures in textureless regions. To fully recognize the objects of various sizes, we design a \textbf{multi-scale} scheme to construct the pairwise relationships in different scopes (i.e. spatial resolutions). Considering the pooling strategy \cite{he2015spatial} might lead to a loss of detail, we adopt the dilation strategy \cite{yu2015multi} to enlarge the receptive field.

In CNNs, low-level features contain rich texture information while high-level features normally reveal semantic clues. With the unary features extracted from different layers, the calculated pairwise relationships are diverse. Therefore, we design a \textbf{multi-level} scheme for LSP. As shown in Figure \ref{model} (a) and (b), under the single-level (SL) setting, we only calculate LSP with the fused convolutional feature. Under the multi-level (ML) setting, the multi-level convolutional features are separately used to calculate LSP.

In practice, $1 \times 1$ convolution is utilized to adjust the channels for each LSP (not shown in the figure). After that, multiple LSPs are fused with the original CFs by the fusion block shown in Figure \ref{model}, which consists of a concatenation operation and one $1 \times 1$ convolutional layer. In this way, the fused feature will learn to balance the appearance information and the structural information by itself.

\subsection{Cost Self-Reassembling}

Stereo matching models tend to make mistakes at occluded areas since the matching candidate points are invisible in the other view. Fortunately, in most cases only parts of the objects are occluded. We can leverage the accurate disparity values obtained in the unoccluded regions to infer those of the occluded regions, according to the disparity relationships among them. This is the basic hypothesis of traditional disparity filling methods \cite{huq2013occlusion}.

Current disparity refinement methods rectify the outliers by predicting the residual with convolutions \cite{khamis2018stereonet, xu2020aanet}. Due to the stationarity of the filter, it is difficult for convolutions to restore various local error patterns. As a result, existing convolution based refinement methods tend to produce over-smooth results to reduce the overall error. To adaptively refine the disparity, we propose CSR, which holds a similar spirit as the disparity filling methods but possesses the advantage of being trainable. 

Specifically, we first excavate the contextual information in the image to understand the disparity relationships between the points. It is realized by a lightweight U-Net \cite{ronneberger2015u} which takes the left image as input and outputs dense features. After that, for each point, several offsets are predicted with the features. Each offset contains two channels, indicating the pixel coordinate of a neighbor point. With the coordinates, we sample the neighbor cost distributions and combine them to update the central one. This reassembling process is formulated as:
\begin{equation}
\mathbf{C}_r(x, y) = \frac{1}{N} \sum_{i=1}^{N} \mathbf{C}_0(x+\Delta x_i, y+\Delta y_i)
\label{equ2}
\end{equation}
where $\mathbf{C}_r$ and $\mathbf{C}_0$ are the refined and initial cost distribution, both with $d_{max}$ channels. They can be converted to the disparity by \emph{softmax} and weighted average. The predicted offset $(\Delta x_i, \Delta y_i)$ might be two decimals and then the sampled values will be obtained by bilinear interpolation. $N$ is a hyperparameter controlling the number of the reassembled neighbors. Intuitively, more points lead to better results but consume more memory.

Considering the contributions of the neighbors might be different, we further introduce the modulation scheme \cite{zhu2019deformable}. For each neighbor point, an additional modulation factor will be predicted to weight the sampled cost. Then Equation \ref{equ2} can be rewritten as:
\begin{equation}
\mathbf{C}_r(x, y) = \frac{1}{\sum_i m_i} \sum_{i=1}^{N} m_i \cdot \mathbf{C}_0(x+\Delta x_i, y+\Delta y_i)
\end{equation}
where $m_i$ is the modulation factor of the $i$-th neighbor point, ranging from $0$ to $1$, output by a \emph{sigmoid} layer.

In addition to the cost distribution, we can apply the dynamic reassembling scheme to the disparity map, as illustrated in Figure \ref{model}. Operating on the disparity, DSR is more similar to the disparity filling methods, formulated as:
\begin{equation}
D_r(x, y) = \frac{1}{\sum_i m_i} \sum_{i=1}^{N} m_i \cdot D_0(x+\Delta x_i, y+\Delta y_i)
\end{equation}
where $D_r$ and $D_0$ are single-channel disparity maps.

Compared with DSR, the advantage of CSR is that it allows us to constrain the refined disparity distribution to be unimodal with the cross entropy loss \cite{tulyakov2018practical}, which has been proved to be effective for the over-smoothing problem \cite{chen2019over}. However, CSR requires more memory since for each assembled neighbor point, we have to store a vector rather than a single value. Although the burden brought by CSR is affordable in our model, we recommend DSR for the designing of practical algorithms.

\subsection{Loss Function}

We use cross entropy loss and smooth $L1$ loss \cite{chang2018pyramid} to respectively supervise the disparity probability distributions and the final disparity results:
\begin{equation}
L_{ce}(\hat{P}(d), P(d)) = \frac{1}{I}\sum_{i=1}^{I} \sum_{d=0}^{d_{max}} -\hat{P_i}(d)\cdot \log P_i(d)
\end{equation}
\begin{equation}
L_{sm}(\hat{D}, D) = \frac{1}{I}\sum_{i=1}^{I} smooth_{L_1}(\hat{D_i}, D_i)
\end{equation}
where $\hat{P}(d)$ is the predicted distribution from \emph{softmax} and $P(d)$ is the ground truth distribution, a normalized Laplacian distribution centered at the disparity ground truth $D$. $\hat{D}$ is the predicted disparity calculated by weighted average. $I$ denotes the number of the pixels in the image.

To sum up, the objective function for training is:
\begin{equation}
L = \sum_{m=1}^{M} \lambda_m(L_{ce}+\mu L_{sm})
\end{equation}
where $M$ is the number of the outputs of our whole model. For GwcNet, we supervise the results from the three hour-glass structures and the one from the refinement module, thus $M$ is 4. $\lambda$ and $\mu$ are the balance weights.

\section{Experiments}

\subsection{Datasets \& Evaluation Metrics}

The utilized datasets and the corresponding evaluation metrics are listed as follows:

\textbf{SceneFlow.} SceneFlow \cite{mayer2016large} is a synthetic dataset containing 35454 training pairs and 4370 testing pairs, both with dense ground truth. We use \textbf{EPE} (End Point Error, calculated by $\frac{1}{\mathcal{N}} \sum_i |\hat{d}_i-d^*_i|$) and \textbf{1-pixel error} (percentage of the points with absolute error larger than 1 pixel) to evaluate the performance on SceneFlow.

\textbf{KITTI.} KITTI datasets consist of KITTI 2012 \cite{geiger2012we} and KITTI 2015 \cite{menze2015object}. The ground truth maps of the training images are sparse and the testing images are without ground truth. On KITTI 2012, there are 194 training pairs and 195 testing pairs. On KITTI 2015, there are 200 training pairs and 200 testing pairs, we split the training pairs into a training set (160) and a validation set (40). The used metrics on KITTI are \textbf{EPE} and \textbf{D1} (percentage of the outliers whose absolute error is larger than 3 pixels and 5\% of the ground truth).

\textbf{Middlebury} \& \textbf{ETH3D.} Middlebury 2014 \cite{scharstein2014high} and ETH3D \cite{schoeps2017cvpr} are two real-world datasets with less than 50 training pairs. Therefore, these two datasets are only used for generalization ability evaluation. \textbf{Bad2.0} (percentage of the points with absolute error larger than 2 pixel) and \textbf{Bad1.0} are the metrics on Middlebury and ETH3D respectively. 

\begin{table*}[t]\small
\centering
\begin{tabular}{lcccccc}
\toprule
	& \multicolumn{2}{c}{Model} & \multicolumn{2}{c}{SceneFlow} & \multicolumn{2}{c}{KITTI 2015}  \\
\cmidrule(l){2-3}
\cmidrule(l){4-7}
	 & {Feature} & {Refinement} & EPE (px) & $>$1px (\%) & EPE (px) & D1-all (\%) 	 \\
\midrule
(a) & CF & \makecell[c]{\XSolidBrush} & 1.04 & 7.68 & 0.68 & 1.65 \\
\midrule
(b) & CF + LSP\_SS & \makecell[c]{\XSolidBrush} & 0.97 & 7.41 & 0.67 & 1.62 \\
(c) & CF + LSP\_SL & \makecell[c]{\XSolidBrush} & 0.99 & 7.55 & 0.67 & 1.68 \\
(d) & CF + LSP\_F & \makecell[c]{\XSolidBrush} & 0.95 & 7.24 & 0.65 & 1.56 \\
\midrule
(e) & CF & ConvNet & 0.84 & 8.05 & 0.63 & 1.72 \\
(f) & CF & DSR	     & 0.80 & 7.03 & 0.61 & 1.49 \\
(g) & CF & CSR	     & 0.79 & 6.70 & 0.61 & 1.35 \\
\midrule
(h) & CF + LSP\_F & CSR & \textbf{0.75} & \textbf{6.52} & \textbf{0.60} & \textbf{1.30} \\
\bottomrule
\end{tabular}
\caption{Quantitative results of the ablation experiments on the SceneFlow test set and KITTI 2015 validation set.}
\label{ablation}
\end{table*}

\subsection{Implementation Details}

The framework is implemented with PyTorch. Adam ($\beta_1=0.9, \beta_2=0.999$) is used as the optimizer. On SceneFlow, we train the model for 10 epochs with a learning rate of $0.001$. On KITTI, we finetune the model trained on SceneFlow for 300 epochs, with the learning rate of 0.001 for the first 200 epochs and 0.0001 for the remainings. For the model submitted to the KITTI benchmark, we use the total training pairs to finetune it for 600 epochs with the learning rate initialized as 0.001 and decayed by 10 times every 200 epochs. At the training phase, images are randomly cropped to $512\times 256$, and the batch size is 4 on two TITAN Xp GPUs.

The maximum disparity is set as 192. In LSP, the neighbor window size is $3\times3$, and the multi-scale dilation rates are 1, 2, 4, 8. In the loss function, $\mu$ is 0.1, the $\lambda$ for the refined result is 1, and others are set as same as in the basic model. 

\begin{figure}[t]
\centering
\includegraphics[width=0.9\columnwidth]{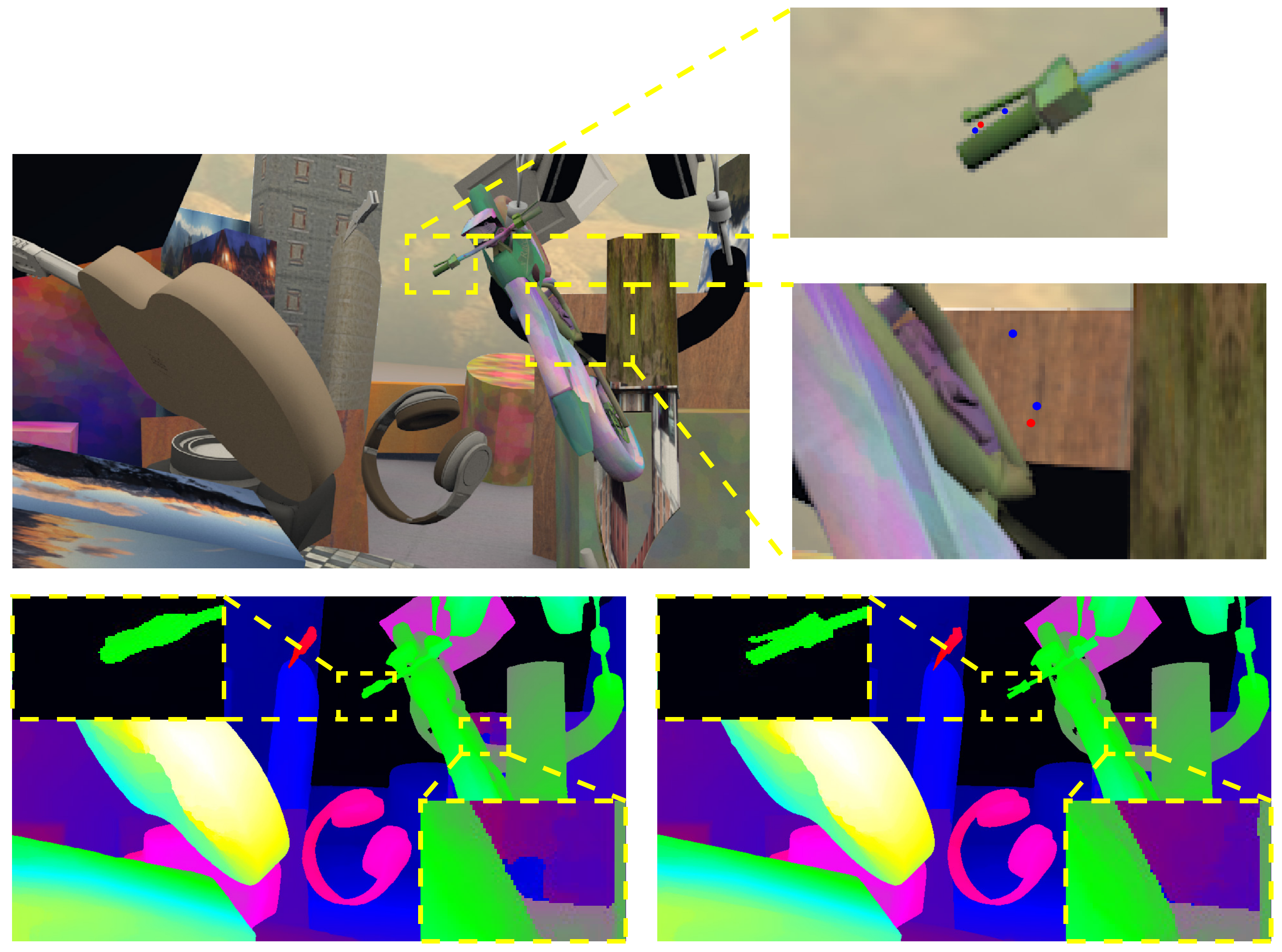} 
\caption{Illustration of the predicted locations in CSR. The red point denotes the central point and the blue ones denote the neighbor points. The disparity results before and after refinement are shown in the bottom.}
\label{fig5}
\end{figure}

\renewcommand{\arraystretch}{1.5}
\begin{table}[t]\small
\centering
\begin{tabular}{l|ccccc}
\hline
$N$ & \textbf{0} & \textbf{1} & \textbf{2} & \textbf{4} & \textbf{8} \\
\hline
EPE (px)   & 0.95 & 0.80 & 0.75 & 0.74 & 0.72  \\ [1pt]
\hline
Memory (G)  & 3.3 & 4.4 & 5.5 & 7.6 & 9.7 \\ [1pt]
\hline
\end{tabular}
\caption{Effects of the assembled point number $N$ in CSR, results are evaluated on SceneFlow.} 
\label{table2}
\end{table}

\begin{figure}[t]
\centering
\includegraphics[width=0.98\columnwidth]{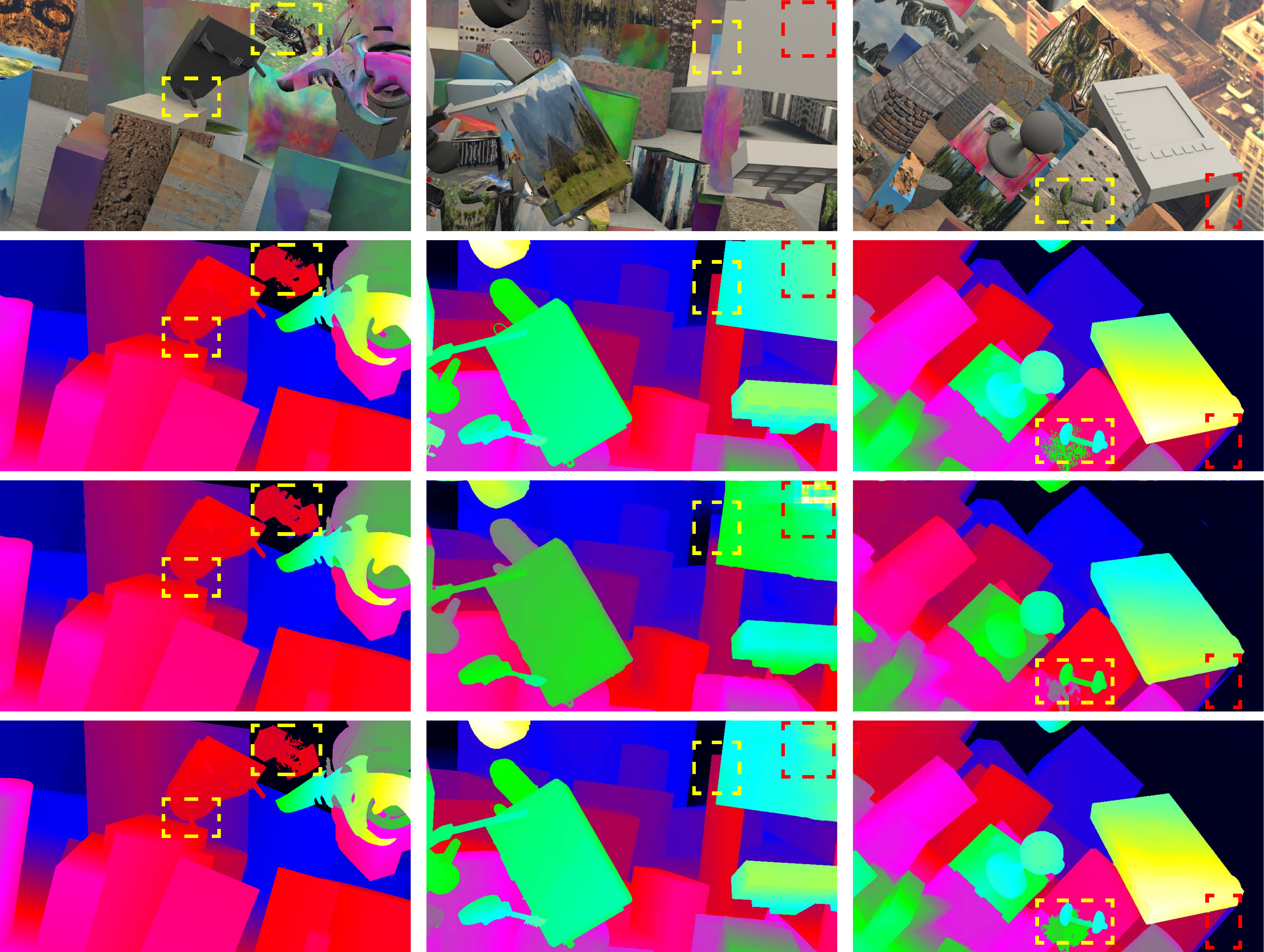} 
\caption{Qualitative results of the ablation experiments on SceneFlow. \textbf{From top to bottom:} input left images, ground truth disparity maps, disparity results from the basic model, and those from our whole model. Please zoom in for details.}
\label{fig4}
\end{figure}

\subsection{Ablation Studies}

The ablation experiments are conducted on SceneFlow and KITTI 2015. GwcNet \cite{Guo_2019_CVPR} with the cost volume formed by feature concatenation is adopted as the basic architecture. The evaluation results are listed in Table \ref{ablation}. Line (a) represents the baseline, where only CF is used to build the cost and no refinement method is utilized. 

In lines (b) to (d), we study the effects of LSP and its multi-scale and multi-level settings. In the table, \emph{SS} denotes single-scale, \emph{SL} represents single-level, and \emph{F} is the full setting. As indicated in the results, fusing LSP with the original CF all achieves a performance increase. We argue that the operation paradigm of convolution makes CF contain less structural information that lies in the pixel relationships. Therefore, LSP is a beneficial complement, especially with our multi-scale and multi-level schemes, which further enriches the structural information. In the following experiments, if there is no special instruction, all LSPs are equipped with the multi-scale and multi-level settings. 

\renewcommand{\arraystretch}{1.5}
\begin{table*}[t]\small
\centering
\begin{tabular}{c|ccccccc}
\hline
Model & PSMNet \shortcite{chang2018pyramid}  & SSPCVNet \shortcite{wu2019semantic} & GwcNet \shortcite{Guo_2019_CVPR} & AcfNet \shortcite{zhang2020adaptive} & StereoDRNet \shortcite{chabra2019stereodrnet} & \\ [1pt]
\hline
EPE (px)   & 1.09   & 0.98 & 0.98 & 0.87 & 0.86  \\ [1pt]
\hline
\hline
Model  & GANet \shortcite{zhang2019ga} & CSPN \shortcite{cheng2019learning} & LEAStereo \shortcite{cheng2020hierarchical} & LaC + GwcNet (ours) & LaC + GANet (ours) \\ [1pt]
\hline
EPE (px)   & 0.80   & 0.78  & 0.78 & 0.75 & \textbf{0.72}   \\ [1pt]
\hline
\end{tabular}
\caption{Evaluation results of current stereo matching algorithms on the SceneFlow test set.} 
\label{benchmark_sf}
\end{table*}

In lines (e) to (g), we compare our DSR and CSR with an existing refinement method \cite{xu2020aanet}. Although EPE can be improved by predicting the disparity residual with a ConvNet, the percentage of the outliers increases as well. This is caused by the over-smoothing problem \cite{chen2019over}. During convolution, an outlier could be corrected but an inlier might also be destroyed since the fixed kernels can not restore various error patterns. On the contrary, CSR and DSR propagate the accurate values in an adaptive manner and are not limited to the regular grid, making them more effective. An example is displayed in Figure \ref{fig5}, as we can see, the predicted neighbors are all belonging to the same semantical object as the central point. Further with the unimodal distribution constraint, CSR improves the \emph{Outlier Percentage} well. In Table \ref{table2}, we evaluate the effects of the assembled point number $N$. Finally considering the tradeoff between the accuracy and memory usage, we set $N$ as 2.

In Table \ref{ablation}, line (h) shows the result of the whole model. Compared with the baseline, we achieve a significant performance increase by inserting the designed modules. Qualitative results on the SceneFlow test set are displayed in Figure \ref{fig4}. As illustrated in the dashed boxes, the disparity in details, textureless regions, and occluded areas are all improved.

\renewcommand{\arraystretch}{1}
\begin{table}[t]\small
\centering
\begin{tabular}{llccc}
\toprule
     & \makecell[l]{Model} & Params. & FLOPs & Memory	 \\
\midrule
(a) & Baseline (B.) 	 & 6.88 M & 1037 G & 3.28 G \\
(b) & B. + LSP 		 & 7.14 M & 1054 G & 3.29 G \\
(c) & B. + LSP + DSR & 9.43 M & 1149 G & 3.45 G \\
(d) & B. + LSP + CSR & 9.43 M & 1150 G & 5.51 G \\
\bottomrule
\end{tabular}
\caption{Complexity analysis of our methods. The resolution of the input images is $960 \times 540$.}
\label{complexity}
\end{table}

%\begin{figure}[t]
%\centering
%\includegraphics[width=0.9\columnwidth]{exp_ic2} 
%\caption{Simulating illumination and color change on SceneFlow. \textbf{First row:} the left image and the disparity ground truth. \textbf{Last three rows from left to right:} the standard or damaged right images, disparity results predicted with CF, and those predicted with LSP.}
%\label{exp_ic}
%\end{figure}

%\begin{table}\small
%\centering
%\begin{tabular}{llccc}
%\toprule
%     & \makecell[l]{Feature} & Standard & w/ IC & w/ CC	 \\
%\midrule
%(a) & CF & \textbf{1.04} & 3.74 & 6.29 \\
%(b) & LSP & 1.08 & \textbf{1.51} & \textbf{2.13}  \\
%\bottomrule
%\end{tabular}
%\caption{Comparisons of the features in different scenes. EPE results on the SceneFlow test set are recorded. \textbf{IC}: illumination change. \textbf{CC}: color change.}
%\label{LSP}
%\end{table}

\subsection{Complexity Analysis}

Since stereo matching algorithms might be deployed to mobile devices, it is important to analyze the complexities. In this section, we record the increase of the parameter number, time consumption, and memory usage after inserting the designed modules. Considering the running time is also related to the used devices, we adopt FLOPs to indicate the time consumption. In this experiment, GwcNet is the basic model and the size of the input images is $960\times540$, i.e. the resolution of the SceneFlow images.

As shown in Table \ref{complexity}, inserting LSP brings little burden for the model. With regard to CSR, while the increases of the FLOPs and parameter number are not significant, it consumes much memory. As discussed before, in CSR, we have to store a vector for each assembled point. On the contrary, the memory consumption of DSR is negligible, making it more suitable for practical algorithms.

\begin{table}\small
\centering
\begin{tabular}{llcc}
\toprule
     & \multirow{2}{*}{Model} & Middlebury  & ETH3D 	 \\
     &				          & Bad2.0-noc (\%)  & Bad1.0-noc (\%) \\	
\midrule
(a) & Baseline(B.) & 22.2 & 9.7  \\
(b) & B. + LSP & 20.1 & 9.6  \\
(c) & B. + LSP + CSR & \textbf{18.7} & \textbf{9.2} \\
\bottomrule
\end{tabular}
\caption{Evaluation results of the generalization ability experiment. Models are trained on SceneFlow.}
\label{generalization}
\end{table}

\begin{figure}[t]
\centering
\includegraphics[width=0.99\columnwidth]{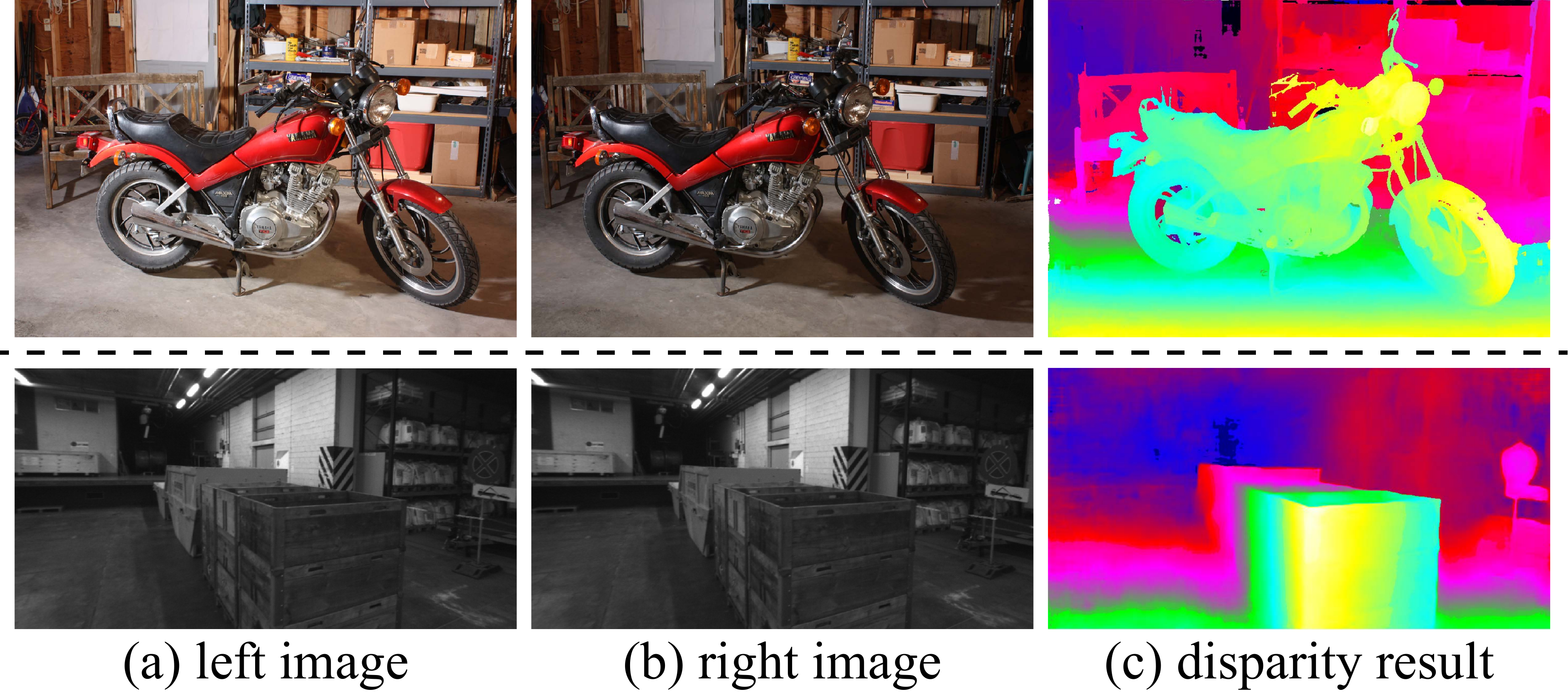} 
\caption{Visualization results of the generalization ability experiment. Images are from Middlebury and ETH3D.}
\label{exp_ga}
\end{figure}

\subsection{Generalization Ability}

In this section, we study the influence of our modules on the generalization ability, with GwcNet as the baseline. Models are trained on SceneFlow and evaluated on the training sets of Middlebury and ETH3D. As presented in Table \ref{generalization}, the designed modules are less affected by the domain gap. On the one hand, CSR avoids leveraging the features of the matching network which are more susceptible than the input image \cite{zhang2020domain}. On the other hand, LSP is beneficial for the illumination inconsistent image pairs from Middlebury, as suggests in \cite{zabih1994non}. Some visualization results of our whole model are displayed in Figure \ref{exp_ga}.

\renewcommand{\arraystretch}{1.2}
\begin{table*}\small
\centering
\begin{tabular}{lccccccccc}
\toprule
\multirow{2}{*}{~~~~~~Model}   & \multicolumn{3}{c}{2015 All (\%)} & \multicolumn{3}{c}{2015 Noc (\%)}   & \multicolumn{2}{c}{2012 All (\%)}  & \multirow{2}{*}{~~~Time (s)}             \\
\cmidrule(l){2-4}
\cmidrule(l){5-7}
\cmidrule(l){8-9}
& D1-bg & D1-fg & D1-all & D1-bg & D1-fg & D1-all & Out-Noc & Out-All \\ 
\midrule
PSMNet \shortcite{chang2018pyramid} 	& 1.86 & 4.62 & 2.32 & 1.71 & 4.31 & 2.14 & 1.49 & 1.89 & 0.41\\
GwcNet-g \shortcite{Guo_2019_CVPR}		& 1.74 & 3.93 & 2.11 & 1.61 & 3.49 & 1.92 & 1.32 & 1.70 & 0.32 \\
AANet+ \shortcite{xu2020aanet}			& 1.65 & 3.96 & 2.03 & 1.49 & 3.66 & 1.85 & 1.55 & 2.04 & 0.06 \\
AcfNet \shortcite{zhang2020adaptive}		& 1.51 & 3.80 & 1.89 & 1.36 & 3.49 & 1.72 & 1.17 & 1.54 & 0.48 \\
CFNet \shortcite{shen2021cfnet}			& 1.54 & 3.56 & 1.88	& 1.43 & 3.25 & 1.73 & 1.23 & 1.58 & 0.18 \\
GANet \shortcite{zhang2019ga}			& 1.48 & 3.46 & 1.81	& 1.60 & 3.11 & 1.63 & 1.19 & 1.60 & 1.80 \\
CSPN \shortcite{cheng2019learning}	 	& 1.51 & 2.88 & 1.74  & 1.40 & 2.67 & 1.61 & - & -	& 1.00 \\
LEAStereo \shortcite{cheng2020hierarchical}	& \textbf{1.40} & 2.91 & \textbf{1.65} & 1.29 & 2.65 & 1.51 & 1.13 & 1.45 & 0.30 \\
\midrule
LaC + GwcNet (ours)  & 1.43 & 3.44 & 1.77 & 1.30 & 3.29 & 1.63 & 1.13 & 1.49 & 0.65 \\
LaC + GANet (ours) & 1.44 & \textbf{2.83} & 1.67 & \textbf{1.26} & \textbf{2.64} & \textbf{1.49} & \textbf{1.05} & \textbf{1.42} & 1.72 \\  
\bottomrule
\end{tabular}
\caption{KITTI online leaderboards, sorted by the \emph{D1-all} of all pixels on KITTI 2015.}
\label{benchmark_kt}
\end{table*}

\begin{figure}[t]
\centering
\includegraphics[width=0.99\columnwidth]{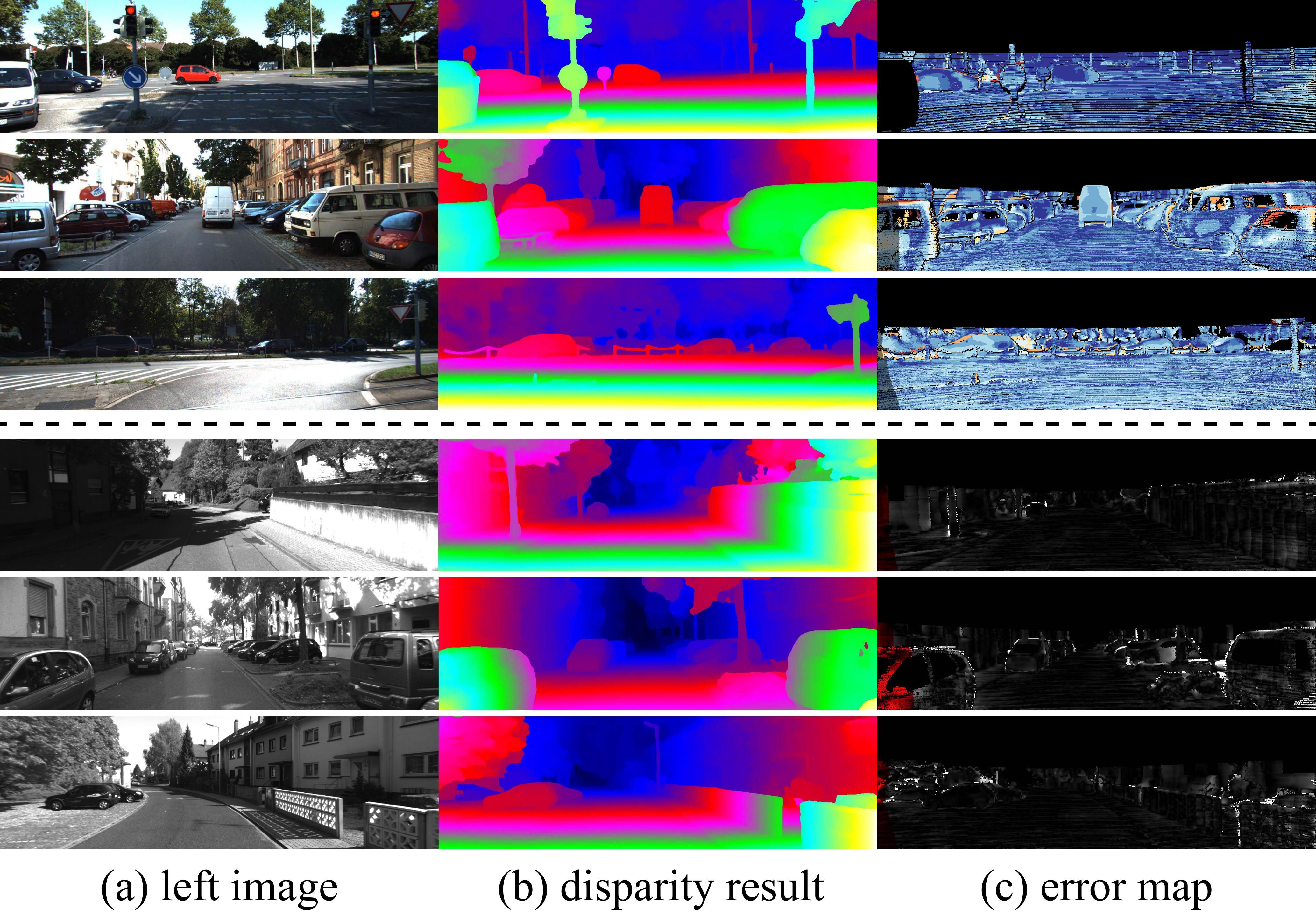} 
\caption{Visualization results of our method on the KITTI test sets. Images in the top three rows are from KITTI 2015 and those in the last three rows are from KITTI 2012.}
\label{kt_fig}
\end{figure}

\subsection{Benchmarks}

In this section, we compare our model with other SOTA algorithms on the SceneFlow dataset and KITTI benchmarks. To further validate the effectiveness, we also implement our modules onto GANet-deep \cite{zhang2019ga}, which is a stronger baseline. 

\textbf{SceneFlow.} For a fair comparison, we only report the methods that share similar training strategies with us. The results are listed in Table \ref{benchmark_sf}, most of them are excerpted from the original papers. GwcNet and GANet are replicated by us with the official codes since there is no result consistent with our training strategy (GwcNet) or the model architecture setting (GANet) in their papers.

\textbf{KITTI.} The evaluation results on the KITTI benchmarks are provided by the online server. In Table \ref{benchmark_kt}, we present some superior methods that only involve the data of the stereo matching task. Several visualization results of our model based on GwcNet are displayed in Figure \ref{kt_fig}.

\textbf{Discussion.} Through inserting the designed modules (LaC, LSP and CSR), we surpass GwcNet by a significant margin on both benchmarks (\textbf{0.23px} on SceneFlow and \textbf{0.34\%} on KITTI 2015). With GANet-deep as the baseline, the increase (\textbf{0.08px} on SceneFlow and \textbf{0.14\%} on KITTI 2015) is not as substantial as GwcNet. In addition to the original performance of GANet is already superior, we analyze the guided aggregation layers in GANet might play a similar role as CSR, i.e. propagating the matching cost based on the image content. Nevertheless, CSR holds the advantage of being able to establish long-range relationships in only one iteration. It is worth mentioning that CSPN \cite{cheng2019learning} and LEAStereo \cite{cheng2020hierarchical} are not adopted in our work since the former model for the stereo matching task is not open-source and the submitted results of the latter can not be reproduced by us with the searched architecture. 

In this paper, we mainly demonstrate the effectiveness of the designed modules in this paper. For real-time applications, we can adopt a faster basic network and incorporate the proposed modules, which will improve the performance with sacrificing a little speed, as shown in Table \ref{complexity}.

\section{Conclusions}

In this paper, we have investigated some limitations of current CNN based stereo matching architectures. In the feature extraction module, aiming at tackling the issue that CF contains less structural information, we design a pairwise feature LSP, which can be combined with CF to enhance the feature representation for more accurate matching. In the disparity refinement module, to mitigate the problem that static convolutions tend to produce over-smooth results, we present an adaptive refinement scheme CSR. CSR is powerful and alleviates the over-smoothing problem well with the unimodal distribution constraint. The practical version of CSR, DSR is also introduced. Experimental results on the SceneFlow and KITTI benchmarks show that the proposed modules can improve the model performance by a significant margin. We confirm that injecting some traditional knowledge into deep networks would be beneficial and meaningful, as explored in previous researches \cite{zhang2019ga, duggal2019deeppruner}.

Future works mainly focus on two aspects: 1) As a novel feature descriptor, LSP has more application scenes worth discovering, such as those with drastic illumination change \cite{zabih1994non}. 2) DSR will be extended to other pixel-level tasks, e.g. semantic segmentation.

% Using the \centering command instead of \begin{center} ... \end{center} will save space
% Positioning your figure at the top of the page will save space and make the paper more readable
% Using 0.95\columnwidth in conjunction with the

% Use \bibliography{yourbibfile} instead or the References section will not appear in your paper
\bibliography{paper_bib}

\end{document}